\documentclass[letterpaper, 10 pt, conference]{ieeeconf}  

\IEEEoverridecommandlockouts                              
                                                          
\usepackage{graphicx}
\usepackage{float}
\usepackage{algorithm}
\usepackage{algpseudocode}
\usepackage{listings}
\usepackage{amsmath}
\usepackage{comment}
\usepackage{hyperref}
\hypersetup{
    colorlinks=true,
    linkcolor=blue,
    filecolor=magenta,      
    urlcolor=cyan,
    pdftitle={Overleaf Example},
    pdfpagemode=FullScreen,
    }
\DeclareUnicodeCharacter{202F}{FIX ME!!!!}

\title{\LARGE \bf
Region Prediction for Efficient Robot Localization on Large Maps
}

\author{Matteo Scucchia$^{1}$ and Davide Maltoni$^{2}$
\thanks{$^{1, 2}$Department of Computer Science and Engineering, University of Bologna, Italy.}%
\thanks{Corresponding author:  {\tt\small matteo.scucchia2@unibo.it}}
}

\begin{document}

\maketitle
\thispagestyle{empty}
\pagestyle{empty}

\begin{abstract}

Recognizing already explored places (a.k.a. place recognition) is a fundamental task in Simultaneous Localization and Mapping (SLAM) to enable robot relocalization and loop closure detection. In topological SLAM the recognition takes place by comparing a signature (or feature vector) associated to the current node with the signatures of the nodes in the known map. However, as the number of nodes increases, matching the current node signature against all the existing ones becomes inefficient and thwarts real-time navigation. In this paper we propose a novel approach to pre-select a subset of map nodes for place recognition. The map nodes are clustered during exploration and each cluster is associated with a region. The region labels become the prediction targets of a deep neural network and, during navigation, only the nodes associated with the regions predicted with high probability are considered for matching. While the proposed technique can be integrated in different SLAM approaches, in this work we describe an effective integration with RTAB-Map (a popular framework for real-time topological SLAM) which allowed us to design and run several experiments to demonstrate its effectiveness. All the code and material from the experiments will be available online at \href{https://github.com/MI-BioLab/region-learner}{https://github.com/MI-BioLab/region-learner}.

\end{abstract}

\section{Introduction}  \label{introduction}

The detection of loop closure, or the determination of whether an agent has returned to a previously visited location by analyzing sensor data, is a crucial aspect of the Simultaneous Localization And Mapping (SLAM) problem because it can significantly improve the accuracy of the map \cite{b1}. Due to intrinsic noise and unmodeled dynamics in its sensors and actuators, as a mobile robot explores its environment, it accumulates a localization error known as drift. By returning to a previously visited location this error can be corrected, resulting in an increase in the overall consistency of the map. From a philosophical point of view, place recognition provides robots with topological knowledge of the world, which is not a never-ending corridor, but rather a space in which the same place can be reached through different paths. Therefore, loop closure detection through effective place recognition techniques is essential for mobile robotic systems capable of long-term autonomous navigation. Unfortunately, this problem is inherently complex and presents significant challenges \cite{b2}; in particular:
\begin{itemize}
    \item variations in environmental conditions, such as differences in lighting between day and night or between different seasons, as well as changes/occlusions resulting from human interactions, can make place recognition difficult.  
    \item to recognize previously visited locations, a robot must maintain a memory of prior observations and compare them with current observations. For large maps, the memory space and processing time required for this process may exceed available resources. For example, the authors of \cite{b3} indicate that a maximum of about 500 nodes can be maintained in memory for real-time loop closure detection during navigation.  
\end{itemize}

To overcome the second challenge, intelligent memory management systems can be adopted, which limit the number of comparisons performed for loop closure detection. For example, RTAB-Map \cite{b4} is a well-known approach that implements a graph-based SLAM with an appearance-based loop closure detector. To ensure real-time performance, RTAB-Map uses three types of memory: 

\begin{enumerate}
    \item Short Term Memory (STM): a buffer which contains the newest nodes of the graph; 
    \item Working Memory (WM): a working area for graph optimization and loop closure detection; 
    \item Long Term Memory (LTM): a local database storing all the knowledge about the map. 
\end{enumerate}

\begin{figure}[htbp]
    \centering
    \includegraphics[width=0.48\textwidth]{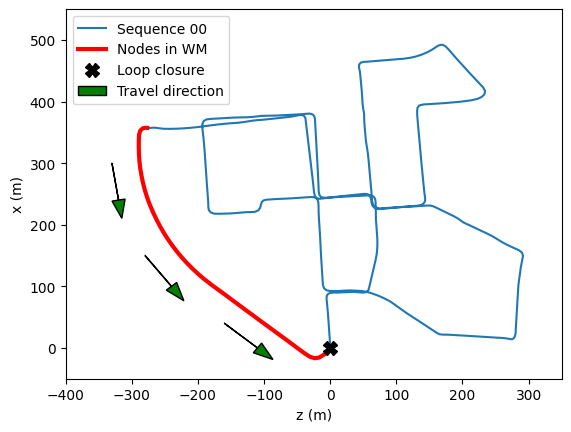}
    \caption{In this example extracted from the sequence 00 of the KITTI odometry dataset, the robot is navigating the path denoted by the green arrows, and WM is populated with the nodes marked by the red trace according to spatio-temporal continuity criteria. Here, the loop closure cannot be detected when the robot crosses the previously explored location, denoted by the symbol x.}
    \label{fig:loop_avoided}
\end{figure}

Thus, only the WM nodes are used for place recognition and loop closure detection. A continuous transfer takes place between WM and LTM in order to: i) retrieve from LTM nodes that are neighbors in time and space with respect to the (recent) highest loop closure hypothesis in WM; ii) transfer back to LTM the nodes that are the less likely to be involved in loop closures, allowing to keep the size of WM within a predefined capacity. Such memory management guarantees a constant time for loop closure detection, regardless of the size of the map. However, if the correct nodes are not present in WM because neither spatial nor temporal continuity is met, detection is unavoidably missed. This is the typical case of a large loop, where the robot returns to an already visited place (see Fig. \ref{fig:loop_avoided}).

\begin{figure*}[htbp]
\centering
\includegraphics[width=0.98\textwidth]{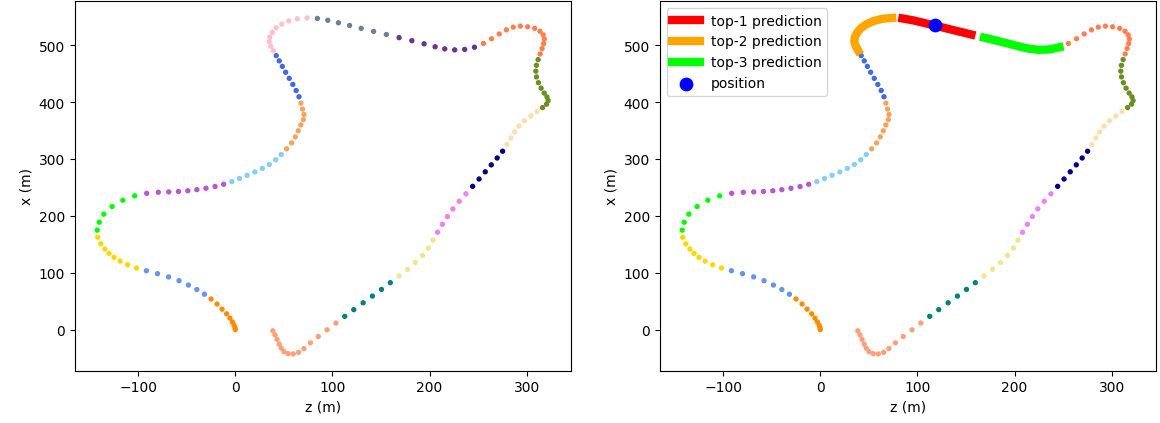}
\caption{The left image shows how the algorithm clusters the graph created by RTAB-Map run on sequence 09 of the KITTI odometry dataset. Each point in the image corresponds to a node in the graph and different clusters are colored differently. The image on the right highlights the regions predicted by the trained model in a subsequent navigation of the same sequence.}
\label{fig:regions}
\end{figure*}

The aim of the proposed method is to ensure that the nodes required for loop closures are always considered. To this purpose, clustering of the map graph is performed during exploration, associating the map nodes to clusters (or regions); each region then becomes the target of a prediction process performed on-line during navigation. In particular, a deep neural network is trained to predict the probability that the current node belongs to the known regions, allowing the system to retrieve all the corresponding nodes. The concept of node clustering is crucial in our approach because signatures (i.e., images in our case) associated to single nodes are too specific and not stable w.r.t. the variations that may occur when the robot revisits the same location, while the set of images associated to a region (e.g., a room in a building or a square in a city) can collectively describe that place in a much more robust way. An example is shown in Fig. \ref{fig:regions}.

It is worth noting that the proposed technique can work with different SLAM approaches, independently of the algorithms used for pose estimation and map optimization. In this paper, we integrated it into RTAB-Map to demonstrate its ability to improve loop closure detection and relocalization in large environments, while still adhering to real-time constraints. No alterations were made to RTAB-Map except for the memory management module. Loop closures are still detected using the RTAB-Map native BoW approach \cite{b5}  and the entire process is still performed using traditional graph-based SLAM (such as pose estimation by PnP Ransac \cite{b6}); our map clustering and deep learning-based region prediction are exclusively used to improve the WM – LTM node exchange. 

After a discussion of the related literature (Section 2), in the rest of the paper our method is explained in detail. Section 3 illustrates the dynamic clustering algorithm used to partition the graph nodes into regions, Sections 4 and 5 detail the region prediction and implementation in RTAB-Map, respectively. In Section 6 we present the experimental results and, finally, in Section 7 we provide some concluding remarks.

\section{Related Works} \label{relatedworks}
Effective memory management and robust place recognition serve as the fundamental tenets for the successful implementation of long-term loop closure detection. A few methods have been proposed in the literature for intelligently managing memory when exploring large environments. One approach, as outlined in \cite{b7}, suggests dividing the exploration of the map into smaller sub-maps, with the creation of a new sub-map automatically triggered upon the passage through a door. This modular structure allows for the independent maintenance of the map for each individual room or outdoor area, facilitating optimization and loop closure detection by considering only a selected portion of the entire map. However, switching between submaps can be complex in non-structured environments. 

\begin{figure}[htbp]
    \centering
    \includegraphics[width=0.35\textwidth]{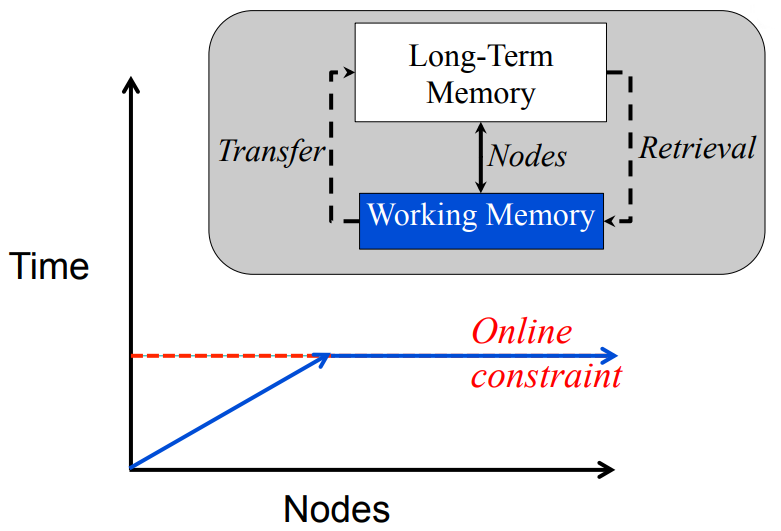}
    \caption{The RTAB-Map computational cost of loop closure detection. The time grows linearly with the number of nodes in the map graph, until the WM is full. Then, the computational cost becomes constant, because always the same number of nodes are kept in the WM.}
    \label{fig:rtabmap}
\end{figure}

RTAB-Map employs three distinct memory structures, as outlined in Section \ref{introduction}. These memories are initially emptied. During robot exploration, when a new node is created, it is inserted into the STM and added to the overall map graph. When the STM reaches its maximum capacity, the least recent nodes are transferred to the WM, where the most recent and most similar nodes w.r.t. the current observation are retained. As the map continues to expand, the number of comparisons for loop closure detection may impede real-time processing. Therefore, a continuous node movement between the WM and LTM is performed (see Fig. \ref{fig:rtabmap}). This process, known as \textit{transfer}, involves transferring the least recent or least similar nodes w.r.t. the new ones in the WM to the LTM. On the contrary, the \textit{retrieval} process involves retrieving nodes from the LTM that are closest in time or space to the current highest loop closure hypothesis and placing them in the WM. As already discussed, and demonstrated in our experimental results, the space and time continuity principles behind the RTAB-Map memory management policy may fail in case of large loop closures.  

With the booming of deep learning, deep neural networks have been utilized to improve place recognition \cite{b8, b9} and continual learning techniques have also been employed to help a robot learn how the appearance of places may change over time \cite{b10}. Novel bio-inspired approaches integrate deep neural networks, continual learning techniques and generative models to improve place recognition \cite{b11}. However, most of these methods still match the current node with the entire map and, therefore, cannot scale to large maps. Therefore, the proposed approach is not an alternative, but can work in conjunction with them to improve efficiency on large maps.  

\section{Map Clustering} \label{clustering}
Map clustering is aimed at partitioning the map nodes in connected clusters (or regions), and in this work it is performed using the scattering-based algorithm proposed in \cite{b12}. This algorithm dynamically creates clusters during exploration, by reducing the spatial \textit{scattering} of the nodes within a cluster. The scattering metric reflects the extent to which the nodes of a cluster are dispersed compared to the equivalent radius. The equivalent radius describes a cluster regardless of its actual shape, comparing it to an ideal high-density topology. How a new node is assigned to a region depends on the desired cardinality of the clusters, a shape factor and the radius upper bound of the equivalent circle. Depending on the environment explored and the speed at which the robot moves, these parameters can be tuned to obtain the desired clustering (avoiding too small or too large regions).

Equation \ref{eqn:scattering} shows how the scattering for a cluster $C_i$ is computed. For a cluster with cardinality $n_i > 1$, $s''$ is the sum of the squared Euclidean distances of all nodes $v$ in the cluster from the centroid $\mathbf{pos}(C_i)$ divided by the equivalent radius $\text{req}(C_i)$ (\ref{eqn:s2}). The threshold $s'$ can be adjusted to regulate the cluster size. See \cite{b12} for more details and examples.

\begin{equation} \label{eqn:scattering}
    s(C_i) = s' + s''(C_i)
\end{equation}
\begin{equation} \label{eqn:s2}
\begin{aligned}
s''(C_i)= \dfrac{1}{\text{req($C_i$)}}\sum_{v\in C_i} \Delta(\mathbf{pos}(v), \mathbf{pos}(C_i))^2 & \text{ if $n_i > 1$}
\end{aligned}
\end{equation}

\begin{figure}[t]
    \centering
    \includegraphics[width=0.48\textwidth]{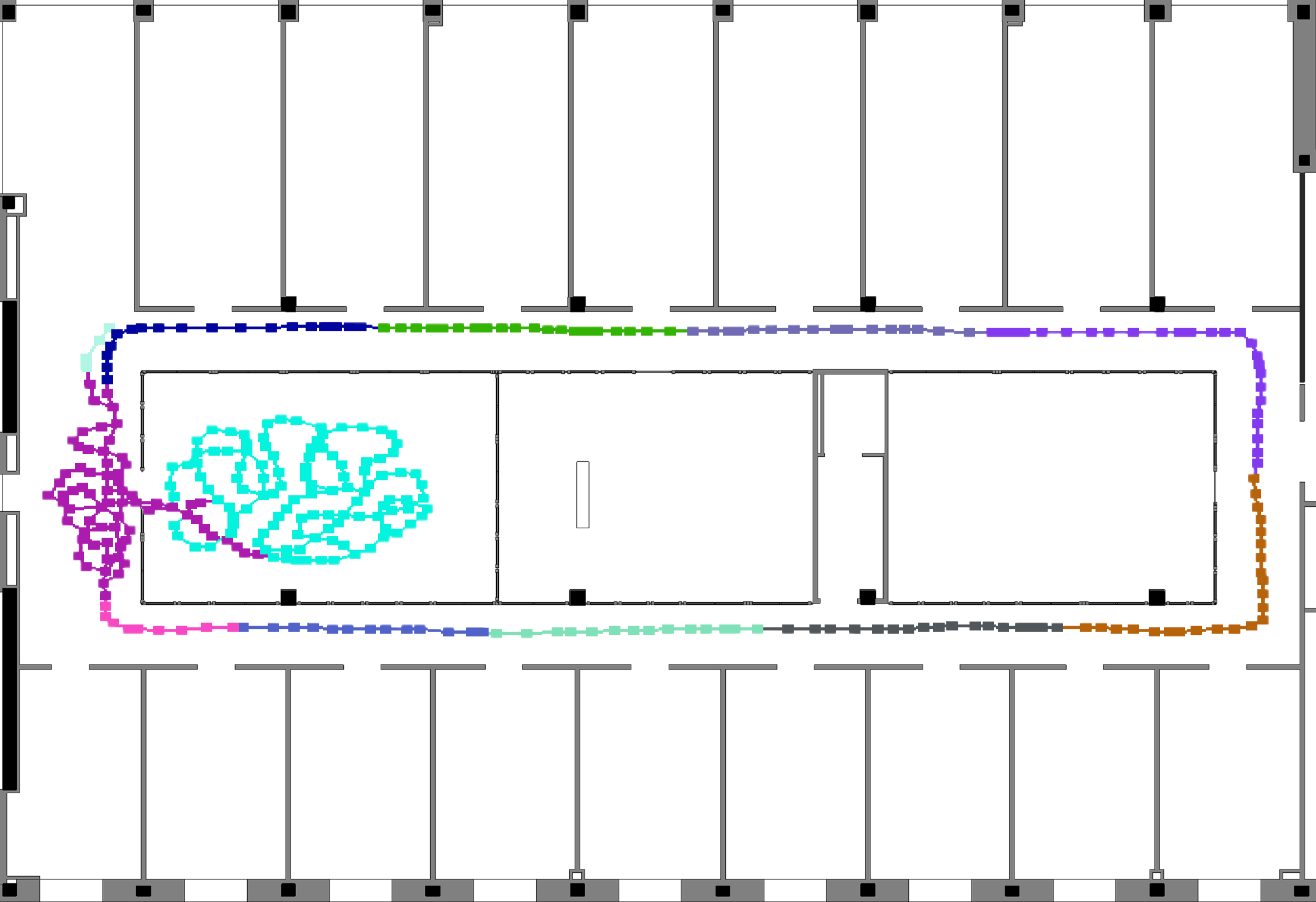}
    \caption{The image shows a portion of one of the floors in our University Campus with overlayed the clustering resulting from a robot navigation. Different clusters/regions are marked by different colors. The majority of the clusters (in the corridors) have a linear shape, however, the clustering algorithm produces two more spherical clusters when the robot explores in deep the room on the left and corridor in front of it, where it closes several small loops triggering node reassignments.}
    \label{fig:biolab}

\end{figure}

Since this approach is a dynamic clustering algorithm, a node can be reassigned to a different cluster if its removal does not violate the connectivity constraint (it does not split the cluster to which it belongs) and if the decrease in scattering caused by its removal in the current cluster is less than the increase in scattering caused by its inclusion in the new cluster. Such a dynamic reassignment allows to create compact (spherical) clusters when the robot moves around in the same region, while without such reassignment cluster shapes would be mostly linear (see Fig. \ref{fig:biolab}).

\section{Region Prediction} \label{prediciton}
While in the general case the neural network used for our region prediction must be trained on-line and in continual learning fashion (i.e., without a clear distinction between training and inference), in this first work, aimed at demonstrating the underlying idea, we adopt some simplifications and identify three stages: 
\begin{enumerate}
    \item the exploration phase in which the robot explores its environment and clusters the map created by RTAB-Map into regions; a training dataset, composed of node images and their region labels, is available at the end of this phase;
    \item the training phase, in which a neural network is trained off-line as a classifier;
    \item the inference phase, in which the neural network is utilized to predict the probabilities of the regions during successive robot’s navigation including new visits of known places.
\end{enumerate}

After the exploration step, the images and zones are used to train the neural network in a supervised manner. A lightweight MobileNetV2 \cite{b13} pre-trained on ImageNet and with frozen weights is used as a convolutional backbone to extract features from images, while a multilayer perceptron (MLP) classifier is trained on the top of the backbone to predict region probabilities. The model used in this work is very simple and allows to perform real-time inference even on embedded computing board. However, according to the desired accuracy/efficiency tradeoff, it can be easily replaced with more complex models tailored to place recognition, such as NetVLAD \cite{b14} and RegionVLAD \cite{b15}.  

\begin{figure}[b]
\centering
\includegraphics[width=0.48\textwidth]{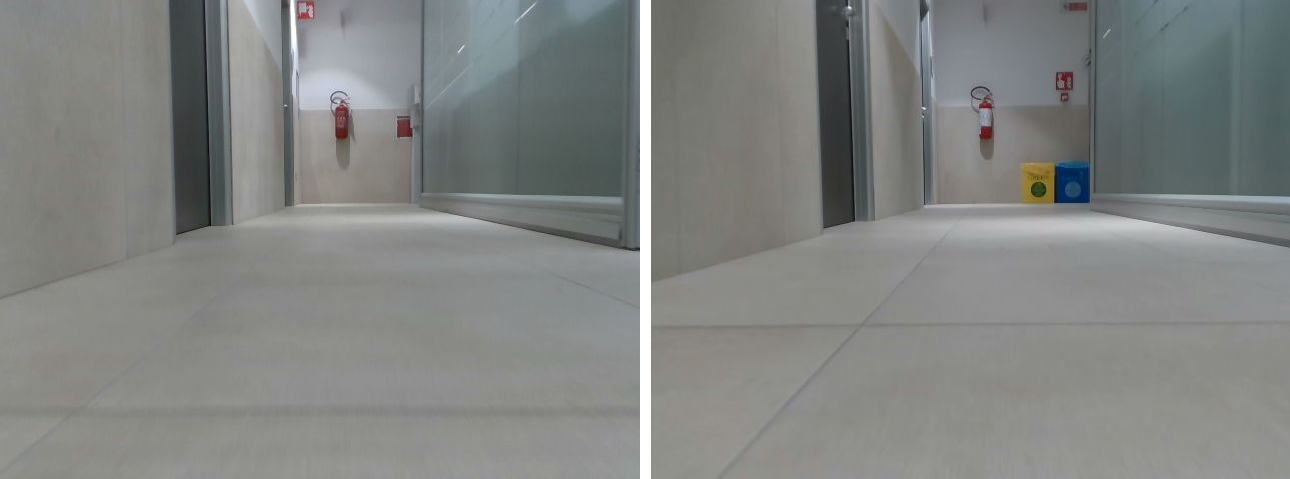}
\caption{Two distinct corridors (from our Campus dataset) whose visual appearance is very similar. In such a case, our model should predict both regions with high confidence.}
\label{fig:corridor}
\end{figure}

The MLP is trained with supervision by assigning a probability of 1 to the correct region and 0 to all other regions. A focal loss is employed, since the cardinality of the clusters may vary significantly, and focal loss better deals with class imbalances. Furthermore, in the last layer of the classifier we replaced the classical soft-max (leading to class probabilities to sum up to 1) with a logistic function to allow the network to predict with high confidence more regions; this is the case of places with scarce textures such as corridors, where the prediction should correctly return more regions (see Fig. \ref{fig:corridor}). At inference time, to mitigate the possibility of retrieving wrong regions (because of high inter-region similarity), an Exponential Moving Average (EMA) was used. The probability $p^i_t$ for the region $i$ at time $t$ depends on the probability of the current observation $o^i_t$ and the probability in the previous step $p^i_{t-1}$ weighted by a parameter $\alpha$. Through this method, the robot may encounter similar environments, but the confidence fusion tends to downweigh the probability of wrong selections. 

\begin{equation} \centering \label{eqn:exp}
    p^i_t = \alpha o^i_t + (1-\alpha)p^i_{t-1}
\end{equation}

{\let\clearpage\relax \subsection{Integration in RTAB-Map}}
The RTAB-Map transfer-retrieval process has been modified to incorporate the retrieval of nodes belonging to the most likely regions during navigation. As in the original solution, WM cannot exceed a maximum size of $N$ nodes. The update procedure (WM\_update in Algorithm \ref{alg:wm_update}) is called every time a new node $v$ is created. 

\begin{algorithm}[] 
\caption{Modified RTAB-Map Transfer-Retrieval}\label{alg:cap}
\begin{algorithmic}[1]
\Procedure{WM\_update}{$v$}
\State $h \gets$ highest loop closure hypothesis of $v$ in WM
\State \parbox[t]{\dimexpr\linewidth-\algorithmicindent}{$U_1 \gets k_1$ neigboring nodes (in time and space) of $h$ in LTM}
\vspace{0.025cm}
\State Retrieve $U_1$
\State $U_2 \gets k_2$ neighboring nodes of $v$ in WM 
\State $O_t \gets$ region predictions of the Neural Network
\State $P_{t-1} \gets$ regions probabilities at previous iteration
\State $P_t \gets \alpha O_t + (1-\alpha)P_{t-1}$  \hfill // Equation \ref{eqn:exp}
\State \parbox[t]{\dimexpr\linewidth-\algorithmicindent}{$U_3 \gets k_3$ node from LTM belonging to the most likely regions according to $P_t$}
\vspace{0.025cm}
\State Retrieve $U_3$
\While{size$(WM) > N$}
\State 
\parbox[t]{\dimexpr\linewidth-\algorithmicindent}{$n \gets$ node in WM belonging to the region \\with lowest probability according to $P_t$ }
\vspace{0.01cm}
\If{$n$ not in $U_1 \cup U_2$} \hfill // Nodes in U1 and U2\\ \hspace{4.9cm} are immunized 
    \State Transfer $n$
\EndIf
\EndWhile
\EndProcedure
\end{algorithmic}
\label{alg:wm_update}

\end{algorithm}

At each step, a set of $k_1$ nodes, which are the neighbors in time and space to the node that turns out to be the highest loop closure hypothesis in WM are retrieved from the LTM. Our method adds to this the retrieval of other $k_3$ nodes belonging to the most likely regions. At this point, the nodes belonging to the less likely regions are transferred back to the LTM, excluding the $k_1$ neighbors retrieved in the current cycle and the $k_2$ neighbors of $v$ already in WM (which are immunized), until WM size is equal to $N$.

\section{Experiments} \label{experiments}

Two types of experiments have been designed to assess the validity of the proposed approach. The former is aimed at estimating the ability of the neural network to predict the correct region when the robot re-visit known places. The latter investigates the accuracy of loop closure detection by comparing the original version of RTAB-Map with the modified one. All the data (including manual and automatic labeling) and code necessary to set up our experiments will be made available for reproducibility and future comparisons.

\begin{figure}[b]
    \centering
    \includegraphics[width=0.3\textwidth]{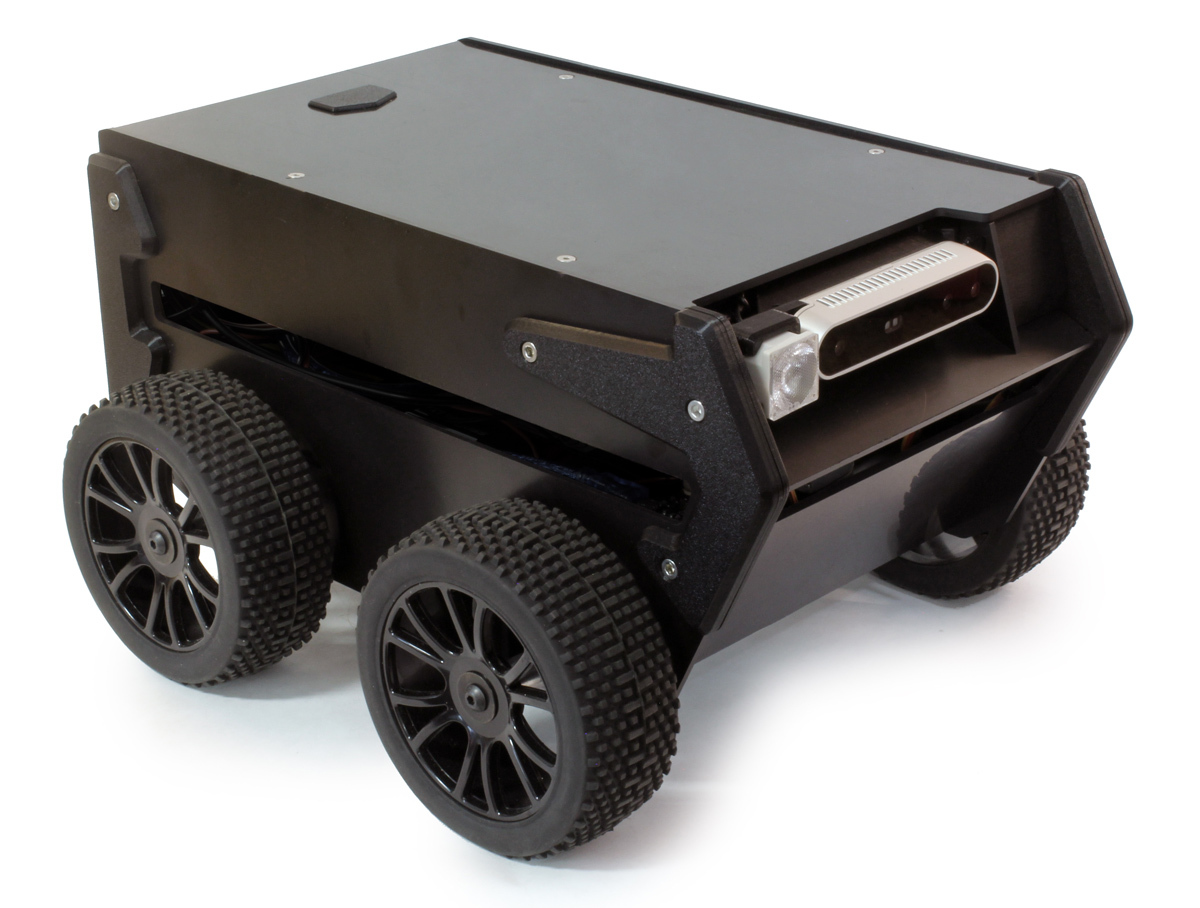}
    \caption{The Xaxxon autocrawler robot employed to acquire the Campus dataset.}
    \label{fig:autocrawler}
\end{figure}

For the experiments, we used three datasets. The first two are well known and widely used in SLAM:\begin{itemize}
    \item \textbf{OpenLoris}: the OpenLoris-Scene dataset \cite{b16}, specifically created for lifelong SLAM, with different indoor sequences with variations in environmental conditions and with the presence of moving people introducing further variability and occlusions.
    \item \textbf{KITTI}: the KITTI odometry dataset \cite{b17}, an outdoor dataset including 22 sequences (11 with ground truth to tune/train the algorithms and 11 without ground truth to test them) acquired with stereo cameras and a 3D velodyne laser in urban and rural settings; 
\end{itemize}
A third dataset (denoted as \textbf{Campus}) was set up using sequences captured with the Xaxxon autocrawler robot \cite{b18} equipped with a 2D rotating lidar and an Intel D435 depth camera (see Fig. \ref{fig:autocrawler}), navigating in three corridors of our university campus. For the experiments performed in this paper, we only used RGB data for region prediction.

\subsection{Prediction Accuracy}
For the first type of experiments, we used Campus and two OpenLoris packages, where the robot navigates the same locations in different sequences: 
\begin{itemize}
    \item the \textbf{Market} package, comprising three sequences captured inside a large marketplace, with relatively consistent lighting conditions across sequences; 
    \item the \textbf{Corridor} package, comprising five sequences captured along a single corridor with large windows, with strongly varying lighting conditions (including artificial illumination turned off).
\end{itemize}

KITTI was not used here because of the lack of sequences revisiting the same locations. For OpenLoris and Campus, the first sequence is used to train the neural network and the other for testing. The region pairing between the training and test sequences was manually identified, by visually matching images and assigning consistent region labels.  

The top-1 and the top-3 error metrics are used for these experiments because they highlight the network’s ability to predict the correct zone and to capture similarities between the different regions.

Table \ref{tab:accuracy} shows the results. In the OpenLoris Market and Campus Corridors sequences, the top-3 accuracy is good, indicating that the correct region is almost always among the three most likely predictions and therefore its nodes have a high chance of being retrieved in WM. Sometimes incorrect regions may be predicted as top-1, however, this behavior is even desirable in some cases, to deal with region similarities. 

\begin{figure}[b]
\centering
\includegraphics[width=0.48\textwidth]{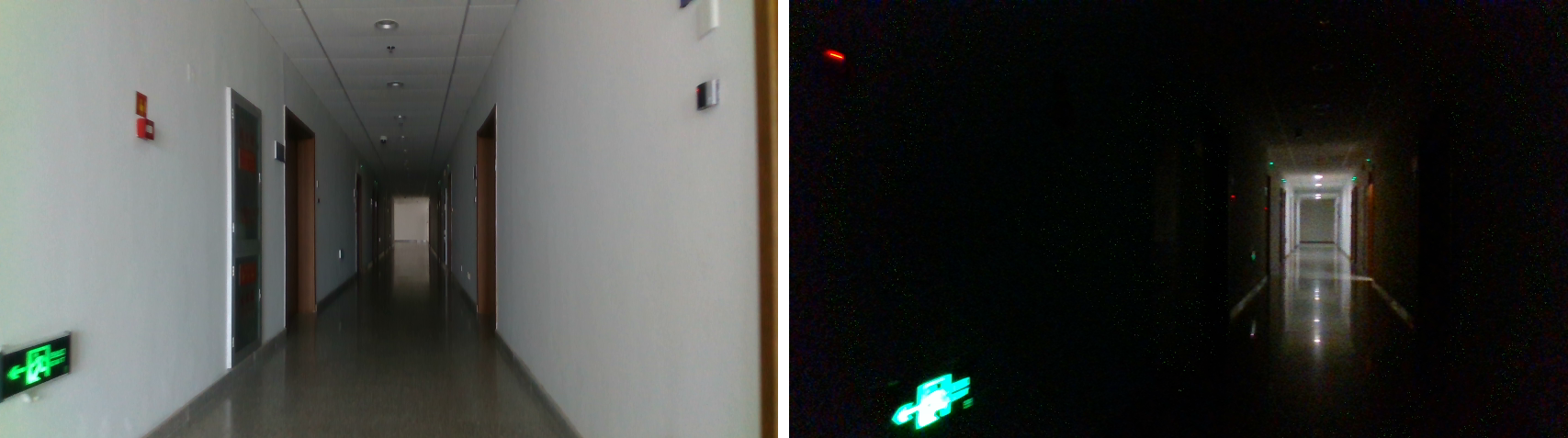}
\caption{On the left is an image from the first Corridor sequence of OpenLoris, with good lighting conditions; on the right is the same corridor in the fourth sequence, with artificial illumination turned off.}
\label{fig:corridor_openloris}
\end{figure}

\begin{table}[ht] 
\caption{Top-1 and Top-3 region prediction accuracy}
\label{tab:accuracy}
\resizebox{0.48\textwidth}{!}{\begin{tabular}{lccc}
\hline
\multicolumn{1}{c}{\textbf{Sequences}} & \textbf{Top-1 accuracy} & \textbf{Top-3 accuracy} \\ \hline
OpenLoris/Market 1-2                  & 0.81           & 0.92           \\
OpenLoris/Market 1-3                   & 0.79           & 0.93           \\
OpenLoris/Corridor 1-2                 & 0.27           & 0.45           \\
OpenLoris/Corridor 1-3                 & 0.06           & 0.14           \\
OpenLoris/Corridor 1-4                 & 0.17           & 0.44           \\
OpenLoris/Corridor 1-5                 & 0.65           & 0.90           \\ 
Campus/Corridor-Biolab 1-2                 & 0.53           & 0.79           \\
Campus/Corridor-DEI 1-2                 & 0.47           & 0.77           \\
Campus/Corridor-Arc 1-2                 & 0.33           & 0.76           \\ \hline
\end{tabular}}
The numbering x-y at the end of each sequence name means that the predictor was trained on sequence x and tested on y.
\vspace{-0.3cm}
\end{table}
The prediction accuracy is not as good on the Corridor sequences of OpenLoris, due to the strong variation in the lighting conditions (see the example in Fig. \ref{fig:corridor_openloris}). To deal with such variation a continual learning of region prediction would be necessary as discussed in Section \ref{conclusions}.

\subsection{Loop Closure Detection}
For the loop closure detection experiments, in addition to the OpenLoris Market package and Campus sequences, we also used KITTI sequences 00-10 that contain loop closures. Referring to the Algorithm \ref{alg:wm_update}, the WM size of RTAB-Map was limited to $N=$ 50 nodes (approximately 15\% of the average size of the sequences considered) to simulate large environment explorations where only a subset of map nodes can be hosted in WM. Parameters $k_1$ and $k_2$ were set to default RTAB-Map values of 2 and 0.25 (i.e. 25\% $ \times$ $N = $ 12 neighboring nodes in WM cannot be transferred to LTM, avoiding the transfer of too recent nodes), respectively. According to the constraints, $k_3 = N - k_1 - k_2$.

Since the ground truth robot pose is known for KITTI and OpenLoris, we implemented a simple algorithm to match poses across the sequences, in order to identify correspondences (and then loop closure candidates) in an automatic way. In particular, if the distance between the (x,y) location in the 2D plane and the angle (also in the 2D plane) between the two poses are less than two given thresholds, then the two poses are considered as a match.

As shown in Table \ref{tab:loop}, our approach was able to recognize all the loop closure in all the sequences considered (except one) totaling 95\% of loop closure detection, even with a very limited WM. On the contrary, the default RTAB-Map memory management was able to detect loop closure only 15\% of times because the spatial and temporal closeness criteria can be unsatisfactory when navigating large maps.

\begin{table}[t] 
\centering
\caption{Loop closure detection results}
\label{tab:loop}
\resizebox{0.48\textwidth}{!}{\begin{tabular}{cccc}
\hline
\textbf{Sequence}          & \textbf{\begin{tabular}[c]{@{}c@{}}N° Loop \\ closures\end{tabular}} & \multicolumn{2}{c}{\textbf{Loops detected}}                                         \\ \hline
\textbf{}                  & \textbf{}                                                            & \textbf{\begin{tabular}[c]{@{}c@{}}RTAB-Map\\ baseline\end{tabular}} & \textbf{Our} \\ \hline
KITTI/00                   & 4                                                                    & 1                                                                    & \textbf{4}   \\
KITTI/02                   & 2                                                                    & 0                                                                    & \textbf{2}   \\
KITTI/05                   & 3                                                                    & 0                                                                    & \textbf{3}   \\
KITTI/06                   & 1                                                                    & 0                                                                    & \textbf{1}   \\
KITTI/07                   & 1                                                                    & 1                                                                    & 1            \\
KITTI/09                   & 1                                                                    & 0                                                                    & \textbf{1}            \\
Campus/Corridor-Biolab 1-2 & 2                                                                    & 0                                                                    & \textbf{2}            \\
Campus/Corridor-DEI 1-2    & 2                                                                    & 0                                                                    & \textbf{2}            \\
Campus/Corridor-Arc 1-2    & 3                                                                    & 1                                                                    & \textbf{2}             \\
OpenLoris/Market 1-3       & 1                                                                    & 0                                                                    & \textbf{1}  \\ 
\hline
Total       & 20                                                                    & 3 (15\%)                                                                    & \textbf{19 (95\%)}  \\ \hline
\end{tabular}}
For Campus and OpenLoris, the numbering x-y at the end of each sequence name means that the predictor was trained on sequence x and tested on y.
\vspace{-0.1cm}
\end{table}

OpenLoris sequences often have no loop closures within the single sequence, but in different sequences the robot returns to the same places and relocalization can be performed (see Table \ref{tab:relocalization}). To deal with these cases, RTAB-Map allows reloading all memory in the WM when a new sequence is started or reloading only the WM as it was at the end of the previous exploration. To keep the WM limited to 50 nodes, the latter option was chosen.

\begin{table}[t]
\centering
\caption{Relocalization results}
\label{tab:relocalization}
\resizebox{0.48\textwidth}{!}{\begin{tabular}{cccc}
\hline
\textbf{Sequence}    & \textbf{N° Relocalization} & \multicolumn{2}{c}{\textbf{\begin{tabular}[c]{@{}c@{}}Relocalizations \\ performed\end{tabular}}} \\ \hline
                     &                             & \textbf{\begin{tabular}[c]{@{}c@{}}RTAB-Map\\ baseline\end{tabular}}        & \textbf{Our}        \\ \hline
OpenLoris/Market 1-2 & 3                           & 1                                                                           & \textbf{3}          \\
OpenLoris/Market 1-3 & 3                           & 1                                                                           & \textbf{3}         \\ 
\hline
Total       & 6                                                                    & 2 (33\%)                                                                    & \textbf{6 (100\%)}  \\ \hline
\end{tabular}}
The numbering x-y at the end of each sequence name means that the predictor was trained on sequence x and tested on y.
\vspace{-0.5cm}
\end{table}

Finally, its is worth noting that even if our node preselection was perfect (i.e., ensuring that the right nodes are always in WM), the native loop closure algorithm of the SLAM used could miss some closures because of insufficient similarity of the corresponding signatures. A posterior analysis of the results showed that the only missed loop by our approach in Table \ref{tab:loop} falls in this category.

\section{Conclusions and Future Works} \label{conclusions}
In this work we introduced a region prediction technique, where map clustering is used to partition nodes into regions and a deep model is trained to predict the region that the robot is currently navigating. This approach enables an efficient implementation of place recognition for real-time loop closure detection. Our experiments validate the effectiveness of the idea but leave some issues open. In fact, while the proposed technique could be already useful in some practical scenario where the predictor is retrained off-line on the whole accumulated map knowledge, its natural evolution is toward an on-line continual learning implementation as required in lifelong SLAM. To this purpose, new regions and new instances (e.g., day/night variations) of already known regions need to be incrementally learned by the model without forgetting the old knowledge (a scenario known as NIC in the continual learning terminology \cite{b19}). Our previous experiences on continual learning make us confident that this target is feasible: in particular, latent replay techniques \cite{b20} seem to be very appropriate in this context because they allow to update the model by storing only a small subset of latent representation of past data.

\end{document}